\pdfoutput=1

\documentclass[11pt]{article}

\usepackage[preprint]{acl}

\usepackage{times}
\usepackage{latexsym}

\usepackage[T1]{fontenc}

\usepackage[utf8]{inputenc}

\usepackage{microtype}

\usepackage{inconsolata}

\usepackage{graphicx}

\usepackage{multirow}
\usepackage{makecell}
\usepackage{courier}
\usepackage{adjustbox}
\usepackage{newfloat}
\usepackage{listings}
\usepackage{colortbl}
\usepackage{booktabs}
\usepackage{pifont}
\usepackage{CJKutf8}
\usepackage[breakable]{tcolorbox}
\definecolor{my_green}{RGB}{40,154,121}
\definecolor{my_red}{RGB}{176,46,46}
\newcommand{\correctmark}{\textcolor{my_green}{\ding{52}}} 
\newcommand{\errormark}{\textcolor{my_red}{\ding{56}}}

\definecolor{YellowGreen}{RGB}{154,205,50}
\definecolor{s_cyan}{RGB}{95,167,177}
\definecolor{s_green}{RGB}{157,181,143}

\newcommand\modelname[0]{\texttt{X-WebAgentBench}}

\newtcolorbox{mybox}[1][]{
	width=\columnwidth,
	colback = gray!4, 
	colframe = black, 
	boxrule = 0.8pt,
  breakable,
	boxsep=0pt,left=10pt,right=10pt,top=8pt,bottom=8pt,
	fontupper=\linespread{1.2}\selectfont,
	title=#1}

\title{X-WebAgentBench: A Multilingual Interactive Web Benchmark for Evaluating Global Agentic System}

\author{
	Peng Wang$^{1,\,2,\,*}$~~
	Ruihan Tao$^{1,}$\thanks{~~Equal Contribution.}~~ 
  	Qiguang Chen$^{1}$~~
  	Mengkang Hu$^{3}$~~
  	Libo Qin$^{1,\,2,}$\thanks{~~Corresponding Author.}
  	\\
	$^{1}$~School of Computer Science and Engineering, Central South University, China \\
	$^{2}$~Key Laboratory of Data Intelligence and Advanced Computing in Provincial Universities,\\Soochow University, China\\
  	$^{3}$~The University of Hong Kong, China \\
	\texttt{wpengxss@gmail.com}, \texttt{aarontao757@gmail.com}, \texttt{lbqin@csu.edu.cn}\\
}

\begin{document}
\maketitle
\begin{abstract}
  Recently, large language model (LLM)-based agents have achieved significant success in interactive environments, attracting significant academic and industrial attention. Despite these advancements, current research predominantly focuses on English scenarios. In reality, there are over 7,000 languages worldwide, all of which demand access to comparable agentic services. Nevertheless, the development of language agents remains inadequate for meeting the diverse requirements of multilingual agentic applications. To fill this gap, we introduce X-WebAgentBench, a novel multilingual agent benchmark in an interactive web environment, which evaluates the planning and interaction performance of language agents across multiple languages, thereby contributing to the advancement of global agent intelligence. Additionally, we assess the performance of various LLMs and cross-lingual alignment methods, examining their effectiveness in enhancing agents. Our findings reveal that even advanced models like GPT-4o, when combined with cross-lingual techniques, fail to achieve satisfactory results. We hope that X-WebAgentBench can serve as a valuable benchmark for multilingual agent scenario in real-world applications.
\end{abstract}

\section{Introduction}

Large language models (LLMs) have demonstrated remarkable success across various natural language processing (NLP) tasks~\cite{zhao2023survey,qin2024large,chen2024unlocking,chen2025towards,10856853}, particularly in the development of language agents~\cite{mu2024embodiedgpt,hu2024treeplanner,hu2024agentgen,shinn2024reflexion}. These agentic systems employ various approaches to interact effectively with the real world. Typically, ReAct~\cite{yao2023react} combines reasoning traces with task-specific actions to mitigate error propagation during chain-of-thought reasoning and interaction, enhancing the reliability of decision-making processes. Furthermore, Reflexion~\cite{shinn2024reflexion} incorporates feedback to induce better interaction. Moreover, MetaGPT~\cite{hong2024metagpt} integrates human workflows to decompose complex tasks in software programming, demonstrating efficiency and scalability in collaborative environments.

\begin{figure}[t]
  \centering
  \includegraphics[width=1.0\columnwidth]{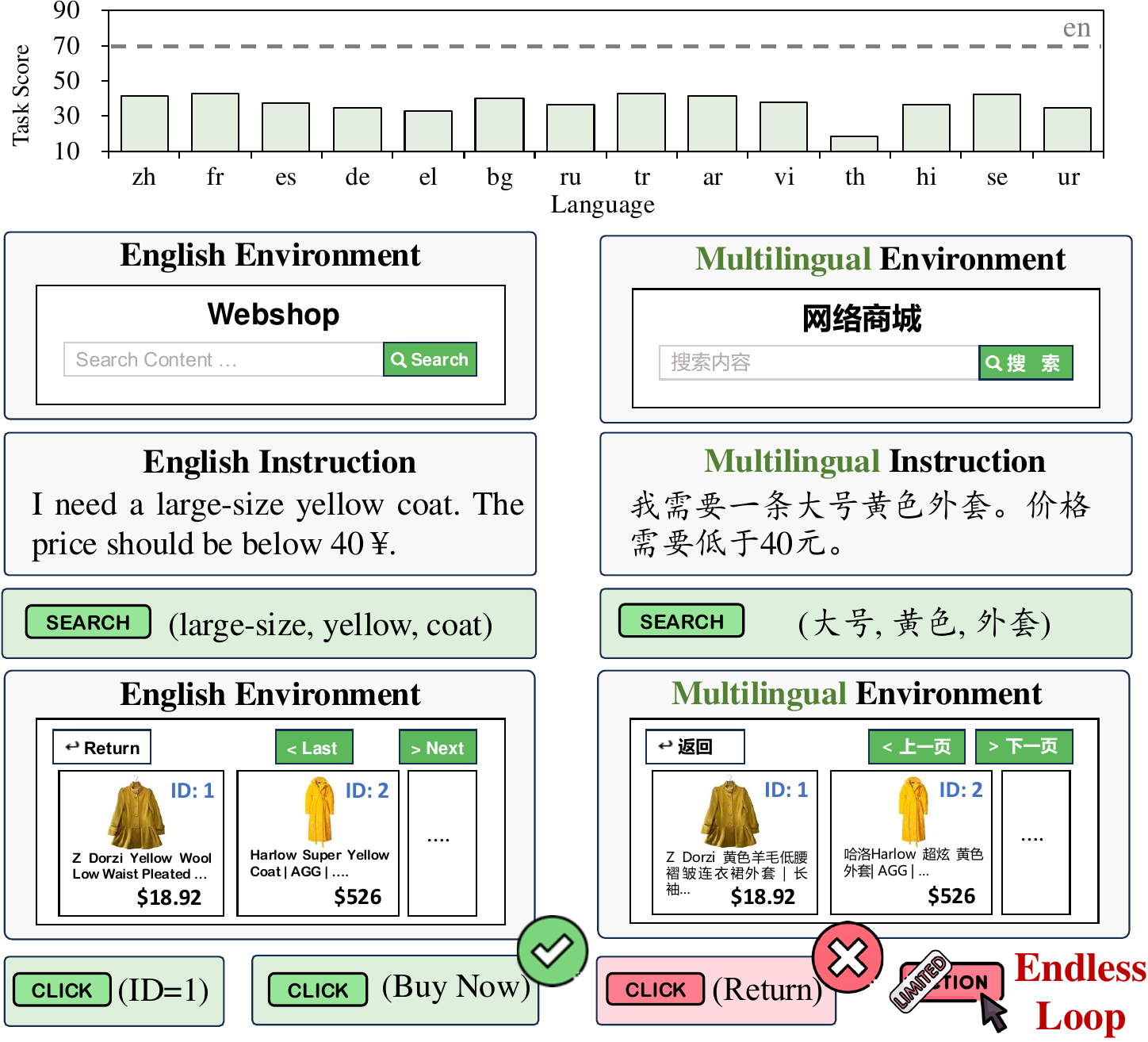}
  \caption{Comparison of performance in English and multilingual settings on GPT-4o: The English task score statistics presented above are derived from \citet{yang2023auto} based on the English WebShop benchmark~\cite{yao2022webshop}, while the multilingual task scores are obtained through evaluation on our own benchmark.}
  \label{fig:intro}
\end{figure}

\begin{table*}[t]
    \centering
    \resizebox{\textwidth}{!}{ 
      \begin{tabular}{lccccc}
        \toprule
        Benchmark & \# Language & \# Tasks & Interactivity & Multilingual Instruction & Multilingual Environment \\
        \midrule
        MiniWoB++~\cite{liu2018reinforcement} & 1     & 100 & \correctmark & \errormark & \errormark \\
        RUSS~\cite{xu-etal-2021-grounding} & 1 & 80 & \correctmark & \errormark & \errormark \\
        WebSRC~\cite{chen-etal-2021-websrc} & 1 & 400k & \correctmark & \errormark & \errormark \\
        WebShop~\cite{yao2022webshop} & 1     & 12,087 & \correctmark & \errormark & \errormark \\
        WebShop-Core~\cite{yao2023react} & 1     & 500 & \correctmark & \errormark & \errormark \\
        Mind2Web~\cite{deng2024mind2web} & 1     & 2,350 & \errormark & \errormark & \errormark \\
        Mind2Web-Live~\cite{pan2024webcanvas} & 1 & 542 & \correctmark & \errormark & \errormark \\
        WebArena~\cite{zhou2024webarena} & 1     & 821   & \correctmark & \errormark & \errormark \\
        OmniACT~\cite{kapoor2025omniact} & 1     & 9,802   & \correctmark & \errormark & \errormark \\
        \midrule
        \modelname{} & 14    & 2,800 & \correctmark & \correctmark & \correctmark \\
        \bottomrule
        \end{tabular}%
    }
    \caption{Comparison of \modelname{} with previous agent benchmarks. \modelname{} could provide multilingual instruction and multilingual interactive environment.}
    \label{tab:compare}%
  \end{table*}%

To better evaluate and understand the capabilities of agents, it emerges a series of research to provide diverse interactive benchmarks to evaluate the ability of language agents in complex environments, such as Mind2Web~\cite{deng2024mind2web}, OsWorld~\cite{xie2024osworld}, and WebArena~\cite{zhou2024webarena}.
Despite the significant advancements in current language agent benchmarks, most developments remain centered on English-language environments. Actually, with globalization accelerating, the real-world application of language agents increasingly involves multilingual contexts~\citep{qin2024multilingual}. For example, as illustrated in Figure~\ref{fig:intro}, users engaging in global e-commerce, such as shopping on Amazon, tend to rely on their native languages for searching, selecting, and purchasing products. Yet, when mainstream language agents are employed in these scenarios, a critical issue emerges: current research often overestimates their effectiveness. Specifically, as illustrated in Figure~\ref{fig:intro}, multilingual performance lags behind English-only performance by over 20\%.
Insights gained from English environments are challenging to generalize to multilingual contexts, which blocks the development of multilingual agents.
Moreover, there is a notable absence of standardized benchmarks for evaluating language agents in multilingual scenario. Without such benchmarks, identifying gaps, limitations, and constraints in existing multilingual agentic systems remains challenging, hindering their global development.

Motivated by this, to fill this gap, we introduce \modelname{}, a multilingual interactive website environment benchmark comprising 14 languages, 2,800 instructions, and 589,946 products. In comparison to previous benchmarks (see Table~\ref{tab:compare}), \modelname{} first offers a comprehensive multilingual setting, with a focus on two primary aspects: (1) \textbf{\textit{multilingual instruction}} (2) \textbf{\textit{multilingual environment}} for language agents. Specifically, \textit{multilingual instruction} aims to evaluate the capabilities to interpret and execute complex commands in various languages based on appropriate actions, ensuring consistent and accurate performance regardless of the linguistic origin of the request.
\textit{Multilingual environment}, on the other hand, focuses on the agent's capability to interact seamlessly with web interfaces and systems designed in multiple languages, guided by the rewards provided by \modelname{}.

Moreover, to investigate the performance and relevant drawbacks of language agents in multilingual scenarios, we conduct comprehensive experiments on \modelname{}, yielding the following key findings: (1) \textit{For larger LLMs, advanced cross-lingual alignment methods can significantly enhance multilingual performance.} (2) \textit{For smaller LLMs, translating multilingual environments into English can effectively mitigate the limitations of their multilingual capabilities.} (3) \textit{The simple integration of existing agent-based and cross-lingual strategies fails to adequately address the challenges posed by multilingual agent systems.}

In summary, our contributions can be summarized as:
\begin{itemize}
    \item We first highlight the research community's overly optimistic estimates of agent systems, as current research primarily focuses on English-centric environments.
    \item To the best of our knowledge, we make the first attempt to introduce a novel multilingual agent benchmark (\modelname{}) that integrates multilingual instructions and environments, taking a meaningful step to build a multilingual agentic system.
    \item Our analysis reveals significant challenges faced by agentic systems in multilingual contexts, particularly in low-resource languages. Further, we provide actionable insights and recommendations to enhance language model performance in these settings.
\end{itemize}

To facilitate the further research, our code will be available at \url{https://github.com/WPENGxs/X-WebAgentBench}.
\begin{figure*}[t]
    \centering
    \includegraphics[width=1.0\textwidth]{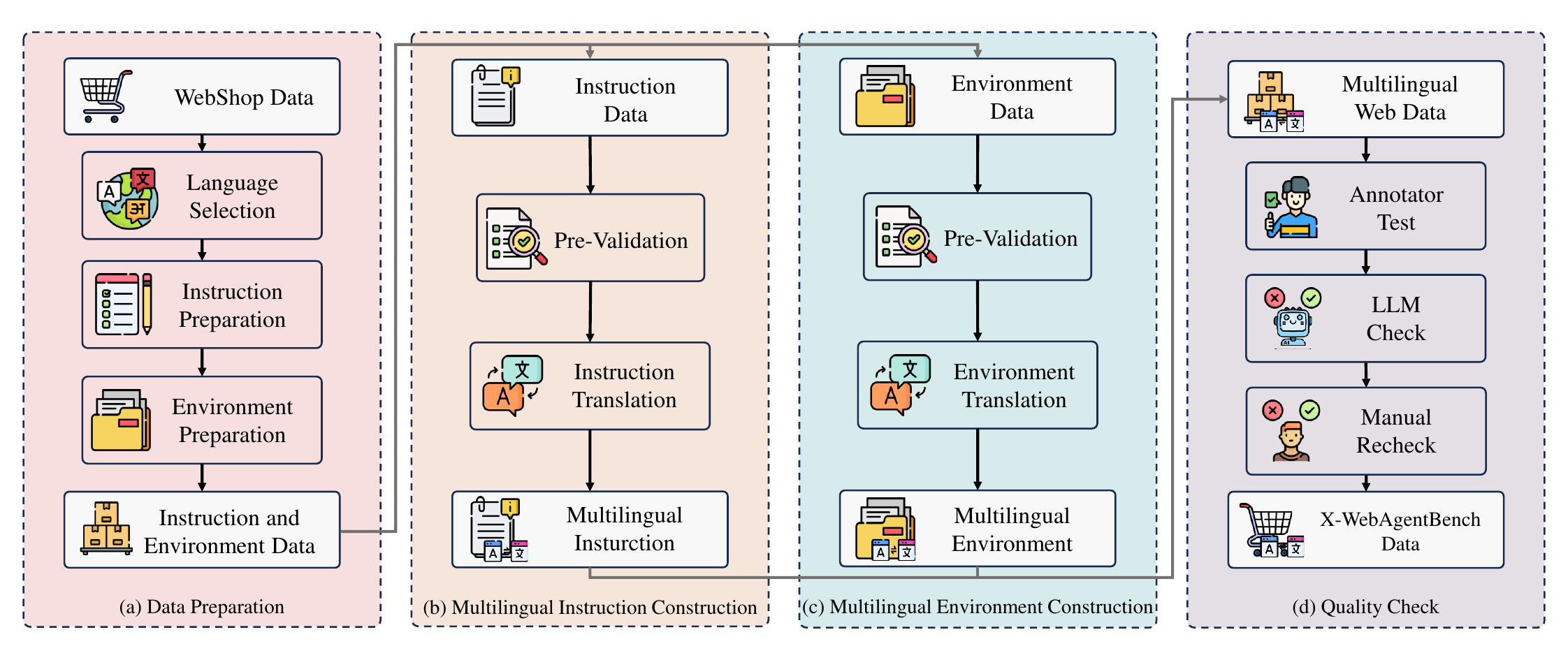}
    \caption{The construction of \modelname{} includes four stages: (a) Data Preparation, (b) Multilingual Instruction Construction, (c) Multilingual Environment Construction, and (d) Quality Check. This workflow figure refers to M$^3$CoT~\cite{chen2024m}.}
    \label{fig:data}
\end{figure*}

\section{\modelname{} Construction}

This section describes the construction of \modelname{}, including Data Preparation (\S~\ref{sec:app:benchmark_detailata-preparation}), Multilingual Instruction Construction (\S~\ref{sec:multilingual-instruction}), Multilingual Envrionment Construction (\S~\ref{sec:multilingual-environment}), and Quality Check (\S~\ref{sec:translation-quality-check}). Further, we will provide the detailed Data Statistics (\S~\ref{sec:statistics}).

\subsection{Data Preparation}
\label{sec:app:benchmark_detailata-preparation}
This section introduces the processing of data preparation, which is illustrated in Figure~\ref{fig:data} (a).

\noindent
\textbf{Language Selection.} Following \citet{conneau2018xnli}, we select 14 representative languages to balance resource distribution and maximize coverage across 7 language families. Moreover, this selection also optimally expands the geographical distribution of spoken languages, thereby increasing its global applicability. The detailed language distribution is shown in Table~\ref{tab:languages} in Appendix~\ref{sec:app:data_detail}.

\noindent
\textbf{Instruction Preparation.} Following the mainstream benchmark setup for language agents~\citep{yao2023react}, we randomly select 500 informative instructions from the WebShop dataset to facilitate product identification. During selection, we manually remove ambiguous instructions, such as ``I need a charger,'' ensuring that each retained instruction corresponds to a well-defined product search.

\noindent
\textbf{Environment Preparation.}
To evaluate the interaction capabilities of LLMs effectively, it is crucial to address the challenges presented by lengthy contexts, particularly in extensive environmental product data. To reduce the impact of long contexts for fair assessment, we simplify the multilingual e-commerce environment from WebShop~\cite{yao2022webshop}. Specifically, each instruction involves an average of 211 products, maintaining a sufficient decision space with appropriate contextual length.

\subsection{Multilingual Instruction Construction}
\label{sec:multilingual-instruction}
This section introduces the construction process of multilingual instruction, which is illustrated in Figure~\ref{fig:data} (b).

\noindent
\textbf{Pre-validation.}
Due to the high cost of manual translation, we explore the use of automated tools for large-scale instruction translation. To assess translation quality, we translated 50 English instructions into 14 languages using automated methods and manually validated the results.
Specifically, we compared two translation tools: prompted GPT-4 and Google Translate. Three experts evaluated whether the translations accurately conveyed the original meaning. The assessment showed that Google Translate consistently achieved accuracy rates above 90\% across all languages. In contrast, GPT-4 performed well for high-resource languages but exhibited significant variability in low-resource languages, with an average accuracy of only 74\%.

\noindent
\textbf{Instruction Translation.} Based on the results in pre-validation, we utilize Google Translate API to autonomously translate these instruction data to the selected 14 languages, which also follows previous multilingual benchmark construction strategies~\citep{conneau2018xnli,zhang2023m3exam}. As a result, we obtain 2,800 multilingual instructions across 14 languages.

\begin{figure*}[t]
    \centering
    \includegraphics[width=1.0\textwidth]{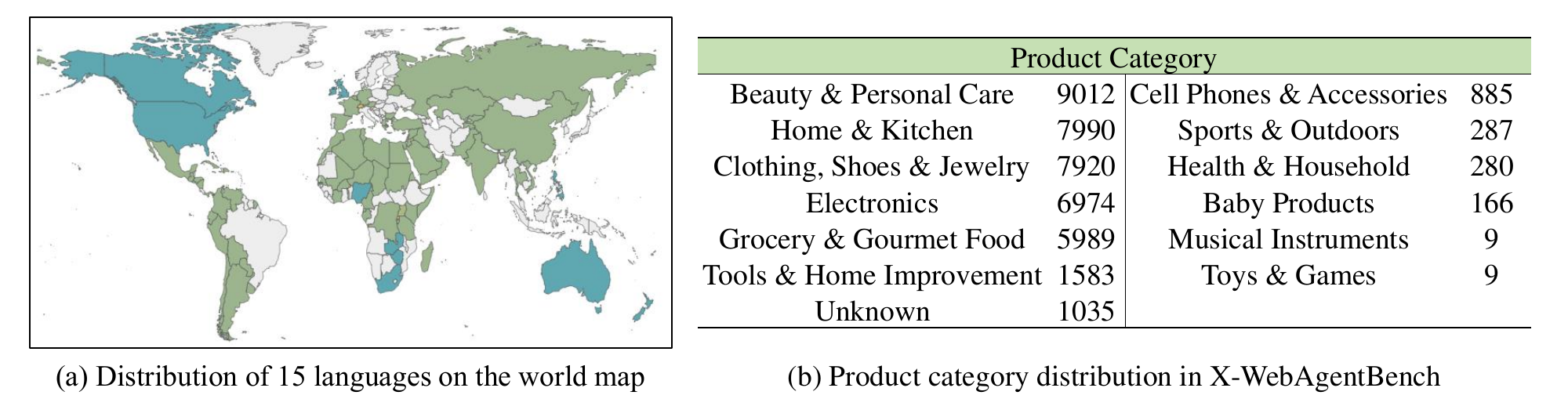}
    \caption{The distribution of languages and product category in \modelname{}, \colorbox{s_cyan}{cyan} represents English area, and \colorbox{s_green}{green} represents multilingual area in \modelname{}.}
    \label{fig:data_statistics}
\end{figure*}
\subsection{Multilingual Environment Construction}
\label{sec:multilingual-environment}

This section, as shown in Figure~\ref{fig:data} (c), presents the construction process of multilingual environment.

\noindent
\textbf{Pre-validation.} Similarly, to ensure data quality in a multilingual environment, we also introduce a pre-validation stage to achieve translation accuracy. We first select 50 products in environment per language to compare GPT-4 and Google Translate. Specifically, we evaluate translations of titles, categories, clickable buttons, product names, full and short descriptions, and customization options. Each field is assessed against corresponding product data for accuracy. Notably, GPT-4 significantly outperforms other tools, even achieving over 15\% higher human annotation accuracy by capturing contextual nuances and producing more precise translations. This step ensures high-quality multilingual data to support the agent's interaction.

\noindent
\textbf{Environment Translation.}
Leveraging pre-validation insights, we standardize all environmental data into a unified JSON format and translate it into 14 languages. In total, we generate 589,946 products to support a robust multilingual environment. Finally, we integrate this product data into an interaction system to construct the multilingual environment in \modelname{}.

\subsection{Quality Check}
\label{sec:translation-quality-check}
Multilingual datasets, created from translations of original English data, always introduce semantic confusion, ambiguity, and cultural differences~\cite{qin2024multilingual,zhang2023m3exam}. For example, in Chinese, both ``trousers'' and ``shorts'' are often represented by the same term, illustrating polysemy issues. To address this, we perform quality checks, as shown in Figure~\ref{fig:data} (d), to ensure data accuracy.

\noindent
\textbf{Onboarding Test.} 
We select 50 test instructions and relevant prodcuts from each of the 14 languages for annotators. To participate in the subsequent recheck task, annotators must achieve an average task score of at least 80 across these items.

\noindent
\textbf{LLM Check.} To ensure semantic accuracy, we adopt the method of \citet{chen2024essential}, comparing translations in multiple languages to the original English using GPT-4. Translations are rated on a 0-to-10 scale, with only those scoring 8 or above being included. This results in approximately 250 instructions and associated products. Notably, most instructions across languages achieve a perfect score of 10, indicating high-quality translations.

\noindent
\textbf{Manual Recheck.} Subsequently, the final version undergoes manual rechecking to verify overall quality, yielding 200 instructions and corresponding products of high quality for each language. A kappa coefficient of 0.8 confirms the reliability of this process~\cite{landis1977measurement}. Details of the recheck process are provided in Appendix~\ref{sec:app:check}.

\subsection{Data Statistics}
\label{sec:statistics}
\modelname{} supports 14 languages, includes 2,800 multilingual text instructions (200 per language), and features 589,946 multilingual products across 13 categories, addressing the linguistic needs of most countries (see Figure~\ref{fig:data_statistics}).
As shown in Table~\ref{tab:action} in Appendix~\ref{sec:app:benchmark_detail}, the multilingual environment in \modelname{} encompass 6 distinct actions, demonstrating a broad decision space.
This variety in multilingual scenarios and agent search spaces enables a comprehensive evaluation of the multilingual and interactive capabilities of LLMs.
\section{Experiments}
\begin{table*}[t]
    \centering
    \resizebox{0.99\textwidth}{!}{ 
        \begin{tabular}{lccccccccccccccc}
        \toprule
        {Method} & {zh} & {fr} & {es} & {de} & {el} & {bg} & {ru} & {tr} & {ar} & {vi} & {th} & {hi} & {sw} & {ur} & {AVG} \\
        \midrule
        \rowcolor[rgb]{ .949,  .949,  .949}  \multicolumn{16}{c}{Mistral-7B-Instruct~\cite{jiang2023mistral}} \\
        \midrule
        \textit{BaseAgent} & \textbf{10.80} & 3.71 & 3.36 & 4.93 & 2.94 & 1.95 & 4.05 & \textbf{3.73} & \textbf{3.02} & 3.06 & 1.88 & 4.01 & 1.31 & 0.33 & 3.51 \\
        \textit{BaseAgent + Translate-en} & 6.08 & \textbf{7.13} & 4.28 & \textbf{5.34} & \textbf{4.25} & \textbf{6.65} & \textbf{5.83} & 3.55 & 1.73 & \textbf{3.95} & \textbf{2.48} & \textbf{6.34} & \textbf{3.35} & \textbf{1.35} & \textbf{4.45} \\
        \textit{BaseAgent + Self-Translate-en} & 1.76 & 3.57 & \textbf{4.32} & 2.53 & 2.12 & 3.48 & 2.06 & 0.20 & 0.29 & 0.98 & 0.49 & 0.71 & 0.00 & 0.00 & 1.61 \\
        \textit{BaseAgent + CLP} & 2.90 & 3.26 & 2.96 & 3.30 & 2.31 & 2.63 & 1.00 & 0.25 & 1.13 & 2.33 & 1.34 & 0.95 & 0.00 & 1.21 & 1.83 \\
        \midrule
        \rowcolor[rgb]{ .949,  .949,  .949} \multicolumn{16}{c}{Llama3-8B~\cite{touvron2023llama}} \\
        \midrule
        \textit{BaseAgent} & 6.35 & \textbf{10.66} & 3.97 & 4.88 & 3.53 & 7.63 & 5.81 & 5.65 & 3.51 & 3.80 & 0.37 & 2.56 & 4.47 & 2.67 & 4.70 \\
        \textit{BaseAgent + Translate-en} & \textbf{11.71} & 9.13 & \textbf{5.44} & \textbf{12.67} & \textbf{5.21} & \textbf{9.77} & \textbf{10.65} & \textbf{8.12} & \textbf{9.02} & \textbf{10.61} & \textbf{1.66} & \textbf{9.52} & \textbf{11.58} & \textbf{4.30} & \textbf{8.53} \\
        \textit{BaseAgent + Self-Translate-en} & 5.04 & 3.30 & 1.79 & 1.79 & 2.21 & 1.75 & 1.99 & 1.97 & 1.34 & 0.83 & 0.45 & 1.47 & 1.21 & 0.50 & 1.83 \\
        \textit{BaseAgent + CLP} & 0.00 & 0.00 & 0.00 & 0.38 & 0.00 & 0.00 & 0.83 & 0.00 & 0.25 & 0.00 & 0.00 & 0.25 & 0.00 & 0.00 & 0.12 \\
        \midrule
        \rowcolor[rgb]{ .949,  .949,  .949} \multicolumn{16}{c}{Qwen2-7B-Instruct~\cite{yang2024qwen2}} \\
        \midrule
        \textit{BaseAgent} & \textbf{27.64} & \textbf{29.47} & \textbf{25.02} & 10.09 & 5.59 & 11.88 & \textbf{25.55} & 14.68 & 9.49 & 14.80 & 6.53 & 10.38 & 10.94 & \textbf{9.07} & 15.08 \\
        \textit{BaseAgent + Translate-en} & 21.37 & 28.04 & 21.30 & \textbf{23.39} & \textbf{22.60} & \textbf{23.63} & 20.35 & \textbf{17.47} & \textbf{13.90} & \textbf{22.62} & 6.35 & \textbf{19.96} & \textbf{13.86} & 7.62 & \textbf{18.75} \\
        \textit{BaseAgent + Self-Translate-en} & 19.31 & 19.32 & 17.43 & 10.98 & 5.93 & 12.51 & 16.33 & 10.30 & 8.62 & 16.09 & \textbf{8.01} & 9.63 & 3.95 & 6.70 & 11.79 \\
        \textit{BaseAgent + CLP} & 15.54 & 8.58 & 2.44 & 1.03 & 7.04 & 9.62 & 10.86 & 4.19 & 6.45 & 6.51 & 0.60 & 1.32 & 1.27 & 8.16 & 5.97 \\
        \midrule
        \rowcolor[rgb]{ .949,  .949,  .949} \multicolumn{16}{c}{GPT-3.5-turbo~\cite{brown2020language}} \\
        \midrule
        \textit{BaseAgent} & 36.04 & 35.40 & 29.61 & 26.06 & 27.12 & 30.17 & \textbf{31.76} & \textbf{29.95} & 28.68 & 23.35 & 11.48 & 12.32 & 33.73 & 15.62 & 26.52 \\
        \textit{BaseAgent + Translate-en} & 27.87 & 24.12 & 13.77 & 28.90 & 29.69 & 31.65 & 30.26 & 16.28 & 14.70 & \textbf{29.89} & 13.40 & \textbf{27.72} & 23.70 & 5.60 & 22.68 \\
        \textit{BaseAgent + Self-Translate-en} & 22.96 & 27.87 & 23.60 & 22.53 & \textbf{34.24} & 23.74 & 22.84 & 26.22 & 12.02 & 28.62 & 12.42 & 24.93 & 26.40 & 8.60 & 22.64 \\
        \textit{BaseAgent + CLP} & \textbf{40.33} & \textbf{39.04} & \textbf{42.81} & \textbf{30.95} & 20.98 & \textbf{34.57} & 22.38 & 27.23 & \textbf{33.41} & 29.53 & \textbf{16.24} & 21.90 & \textbf{34.16} & \textbf{23.58} & \textbf{29.79} \\
        \midrule
        \rowcolor[rgb]{ .949,  .949,  .949} \multicolumn{16}{c}{GPT-4o~\cite{achiam2023gpt}} \\
        \midrule
        \textit{BaseAgent} & \textbf{41.65} & 42.70 & 37.31 & 34.56 & 32.80 & 40.10 & 36.41 & \textbf{43.18} & 41.34 & 37.93 & 18.51 & 36.75 & \textbf{42.10} & 34.15 & 37.11 \\
        \textit{BaseAgent + Translate-en} & 34.80 & 48.33 & 25.97 & \textbf{37.41} & 28.92 & 36.74 & \textbf{38.00} & 33.99 & 2.18 & 35.19 & 12.69 & 33.47 & 36.49 & 2.30 & 29.03 \\
        \textit{BaseAgent + Self-Translate-en} & 20.98 & 33.39 & 27.45 & 18.58 & 26.63 & 29.20 & 29.15 & 12.75 & 8.40 & 26.72 & 18.33 & 10.45 & 26.41 & 20.42 & 22.06 \\
        \textit{BaseAgent + CLP} & 25.75 & \textbf{48.39} & \textbf{42.70} & 36.83 & \textbf{42.24} & \textbf{41.79} & 37.04 & 41.87 & \textbf{41.40} & \textbf{38.44} & \textbf{21.79} & \textbf{37.68} & 40.01 & \textbf{35.18} & \textbf{37.94} \\
        \bottomrule
        \end{tabular}%
    }
    \caption{Main results for \modelname{}. \textbf{Bold number} indicates the best performance for that language in the current model and methods. ``AVG'' presents the average task score in 14 languages.}
    \label{tab:main}%
  \end{table*}%
\subsection{Experimental Setting}
We evaluate 5 LLMs in \modelname{}, including GPT-4o~\cite{achiam2023gpt}, GPT-3.5-turbo~\cite{brown2020language}, Qwen2-7B-Instruct~\cite{yang2024qwen2}, Mistral-7B-Instruct~\cite{jiang2023mistral}, and Llama3-8B~\cite{touvron2023llama}.
Following WebShop~\cite{yao2022webshop}, we use the Task Score to evaluate the performance of language agents.

\subsection{Baselines}
For each LLM we utilize 4 baseline strategies to test the performance in \modelname{}, including: 
(1) \textit{BaseAgent}~\cite{liu2024agentbench} interacts with \modelname{} in local language based on~\citet{liu2024agentbench}.
(2) \textit{Translate-en}~\cite{shi2023language} translates \modelname{} to English with Google Translate.
(3) \textit{Self-Translate-en}~\cite{etxaniz2023multilingual} relies on LLM translating \modelname{} to English.
(4) \textit{CLP}~\cite{qin2023cross} uses cross-lingual alignment prompting to enable \textit{BaseAgent} to understand \modelname{}, followed by task-specific solver prompting for final solution.

\subsection{Results for \modelname{}}
From the results of \modelname{} shown in the Table~\ref{tab:main}, we have the following observations:

\begin{table*}[t]
    \centering
    \begin{adjustbox}{width=\textwidth}
      \begin{tabular}{lcccccccccccccccc}
            \toprule
            Model & Action Type & zh    & fr    & es    & de    & el    & bg    & ru    & tr    & ar    & vi    & th    & hi    & sw    & ur    & AVG \\
            \midrule
            \multirow{3}[2]{*}{\makecell[l]{Qwen2-7B-Instruct\\\cite{yang2024qwen2}}} & All   & \cellcolor[rgb]{ .996,  .965,  .961}5.70  & \cellcolor[rgb]{ .992,  .945,  .941}6.09  & \cellcolor[rgb]{ .996,  .965,  .961}5.73  & \cellcolor[rgb]{ .953,  .671,  .639}11.43  & \cellcolor[rgb]{ .996,  .953,  .949}5.96  & \cellcolor[rgb]{ 1,  .976,  .973}5.50  & \cellcolor[rgb]{ .996,  .965,  .961}5.73  & \cellcolor[rgb]{ .976,  .824,  .804}8.51  & \cellcolor[rgb]{ .996,  .953,  .949}5.93  & \cellcolor[rgb]{ .988,  .898,  .89}7.00  & \cellcolor[rgb]{ .98,  .847,  .831}8.05  & \cellcolor[rgb]{ .988,  .914,  .906}6.71  & \cellcolor[rgb]{ .984,  .886,  .875}7.28  & \cellcolor[rgb]{ .984,  .882,  .875}7.30  & 6.92  \\
                  & Search & \cellcolor[rgb]{ .988,  .894,  .882}1.61  & \cellcolor[rgb]{ .984,  .882,  .871}1.68  & \cellcolor[rgb]{ .996,  .969,  .969}1.18  & \cellcolor[rgb]{ .988,  .91,  .902}1.52  & \cellcolor[rgb]{ .996,  .969,  .965}1.20  & \cellcolor[rgb]{ .996,  .969,  .965}1.20  & \cellcolor[rgb]{ .996,  .953,  .949}1.27  & \cellcolor[rgb]{ .992,  .941,  .937}1.34  & \cellcolor[rgb]{ .992,  .933,  .925}1.40  & \cellcolor[rgb]{ .992,  .925,  .918}1.43  & \cellcolor[rgb]{ .953,  .671,  .639}2.87  & \cellcolor[rgb]{ .988,  .894,  .882}1.62  & \cellcolor[rgb]{ .988,  .898,  .886}1.60  & \cellcolor[rgb]{ .992,  .945,  .937}1.33  & 1.52  \\
                  & Click & \cellcolor[rgb]{ 1,  .996,  .996}4.08  & \cellcolor[rgb]{ 1,  .98,  .976}4.41  & \cellcolor[rgb]{ .996,  .973,  .969}4.55  & \cellcolor[rgb]{ .953,  .671,  .639}9.90  & \cellcolor[rgb]{ .996,  .961,  .957}4.76  & \cellcolor[rgb]{ 1,  .984,  .984}4.30  & \cellcolor[rgb]{ 1,  .976,  .973}4.46  & \cellcolor[rgb]{ .969,  .769,  .745}8.17  & \cellcolor[rgb]{ .996,  .973,  .969}4.54  & \cellcolor[rgb]{ .988,  .914,  .906}5.57  & \cellcolor[rgb]{ .992,  .937,  .929}5.19  & \cellcolor[rgb]{ .988,  .91,  .898}5.68  & \cellcolor[rgb]{ .988,  .91,  .898}5.68  & \cellcolor[rgb]{ .984,  .89,  .882}5.97  & 5.52  \\
            \midrule
            \multirow{3}[2]{*}{\makecell[l]{GPT-3.5-turbo\\\cite{brown2020language}}} & All   & 4.70  & \cellcolor[rgb]{ .98,  .863,  .851}5.86  & \cellcolor[rgb]{ .988,  .894,  .886}5.67  & \cellcolor[rgb]{ .996,  .961,  .957}5.25  & 4.97  & \cellcolor[rgb]{ .988,  .902,  .894}5.62  & \cellcolor[rgb]{ .976,  .82,  .8}6.15  & \cellcolor[rgb]{ .996,  .969,  .965}5.21  & \cellcolor[rgb]{ .988,  .918,  .91}5.53  & \cellcolor[rgb]{ .988,  .898,  .89}5.64  & \cellcolor[rgb]{ .953,  .671,  .639}7.06  & \cellcolor[rgb]{ .957,  .69,  .663}6.94  & \cellcolor[rgb]{ .984,  .871,  .859}5.81  & \cellcolor[rgb]{ .984,  .89,  .878}5.71  & 5.72  \\
                  & Search & \cellcolor[rgb]{ .988,  .906,  .898}1.67  & \cellcolor[rgb]{ .98,  .863,  .851}1.97  & \cellcolor[rgb]{ .992,  .922,  .914}1.57  & \cellcolor[rgb]{ .988,  .898,  .89}1.71  & \cellcolor[rgb]{ .992,  .929,  .922}1.50  & \cellcolor[rgb]{ .988,  .91,  .902}1.64  & \cellcolor[rgb]{ .976,  .812,  .796}2.31  & \cellcolor[rgb]{ .988,  .898,  .886}1.73  & \cellcolor[rgb]{ .992,  .933,  .925}1.49  & \cellcolor[rgb]{ .992,  .922,  .914}1.56  & \cellcolor[rgb]{ .953,  .671,  .639}3.29  & \cellcolor[rgb]{ .961,  .702,  .675}3.09  & \cellcolor[rgb]{ .976,  .812,  .796}2.31  & \cellcolor[rgb]{ .98,  .859,  .847}1.99  & 1.99  \\
                  & Click & 3.02  & 3.89  & \cellcolor[rgb]{ .953,  .671,  .639}4.10  & 3.54  & 3.46  & 3.98  & 3.83  & 3.48  & \cellcolor[rgb]{ .984,  .871,  .859}4.04  & \cellcolor[rgb]{ .965,  .737,  .714}4.08  & 3.77  & 3.85  & 3.50  & 3.72  & 3.73  \\
            \midrule
            \multirow{3}[2]{*}{\makecell[l]{GPT-4o\\\cite{achiam2023gpt}}} & All   & \cellcolor[rgb]{ .996,  .961,  .957}5.37  & \cellcolor[rgb]{ .992,  .933,  .929}5.59  & 4.97  & \cellcolor[rgb]{ .992,  .922,  .914}5.72  & \cellcolor[rgb]{ .996,  .957,  .953}5.38  & \cellcolor[rgb]{ .992,  .937,  .929}5.58  & \cellcolor[rgb]{ .984,  .89,  .878}5.98  & 5.00  & \cellcolor[rgb]{ 1,  .976,  .976}5.21  & \cellcolor[rgb]{ .996,  .957,  .953}5.39  & \cellcolor[rgb]{ .953,  .671,  .639}7.90  & \cellcolor[rgb]{ .988,  .894,  .882}5.95  & \cellcolor[rgb]{ .996,  .961,  .957}5.35  & \cellcolor[rgb]{ .996,  .969,  .965}5.29  & 5.62  \\
                  & Search & \cellcolor[rgb]{ .996,  .961,  .957}1.32  & \cellcolor[rgb]{ .98,  .855,  .843}2.17  & \cellcolor[rgb]{ .992,  .925,  .918}1.61  & \cellcolor[rgb]{ .992,  .922,  .914}1.65  & \cellcolor[rgb]{ .992,  .945,  .941}1.44  & \cellcolor[rgb]{ .996,  .953,  .949}1.39  & \cellcolor[rgb]{ .976,  .827,  .812}2.39  & \cellcolor[rgb]{ .992,  .945,  .941}1.45  & \cellcolor[rgb]{ .996,  .957,  .953}1.35  & \cellcolor[rgb]{ .996,  .949,  .945}1.42  & \cellcolor[rgb]{ .953,  .671,  .639}3.63  & \cellcolor[rgb]{ .976,  .827,  .812}2.38  & \cellcolor[rgb]{ .988,  .902,  .894}1.79  & \cellcolor[rgb]{ .988,  .918,  .91}1.68  & 1.83  \\
                  & Click & \cellcolor[rgb]{ .996,  .953,  .945}4.04  & 3.42  & 3.36  & \cellcolor[rgb]{ .988,  .902,  .89}4.08  & 3.95  & \cellcolor[rgb]{ .965,  .749,  .725}4.20  & 3.59  & 3.55  & 3.85  & 3.97  & \cellcolor[rgb]{ .953,  .671,  .639}4.26  & 3.56  & 3.56  & 3.61  & 3.79  \\
            \bottomrule
            \end{tabular}%
    \end{adjustbox}
    \caption{The average action steps of different language in Qwen2-7B-Instruct, GPT-3.5-turbo, and GPT-4o. We fill color scales when all actions are greater than 5, search actions are greater than 1, and click actions are greater than 4. The deeper color means the more actions compared to other languages.}
    \label{tab:step}%
  \end{table*}%

\noindent\textbf{\textit{(1) Cross-lingual self-alignment is only effective for LLMs with advanced multilingual capabilities.}}
While original cross-lingual self-alignment methods are applicable to smaller LLMs~\citep{etxaniz2023multilingual,qin2023cross}, our findings show that only LLMs with advanced multilingual capabilities~\cite{qin2024multilingual}, such as GPT-series, enable self-alignment techniques like \textit{CLP} to outperform other baselines in multilingual settings. In contrast, weak open-source LLMs consistently fail. We attribute it to the limited multilingual capacity in adapting to complex interactive environments of weak LLMs, which prevents cross-lingual self-alignment from functioning properly, often leading to significant performance degradation.

\noindent\textbf{\textit{(2) Smaller LLMs ($\leq$ 8B) need external Translation tools for effective cross-lingual alignment.}}
As shown in Table~\ref{tab:main}, open-source LLMs with fewer than 8B parameters underperform in multilingual settings using \textit{BaseAgent} and other self-alignment strategies. Only methods that translate languages into English using external translation tools, significantly enhance cross-lingual alignment, leading to task score improvements in \modelname{}.

\noindent\textbf{\textit{(3) Multilingual interactive scenarios remain a significant challenge for current agentic systems.}}
Even GPT-4o, when integrated with advanced \textit{CLP} strategies, fails to produce satisfactory results. This indicates that the persistent performance gap between multilingual and English environments in language agentic systems. The decrease in performance due to the language gap renders most current agent methods nearly unusable.

\section{Analysis}
In this section, we analyze the findings from both the multilingual(\S~\ref{sec:token} and \S~\ref{sec:language_steps}) and agentic perspectives(\S~\ref{sec:app:data_detailction-step} and Appendix \S~\ref{sec:app:error}) as follows:

\begin{figure}[t]
    \centering
    \includegraphics[width=1.0\columnwidth]{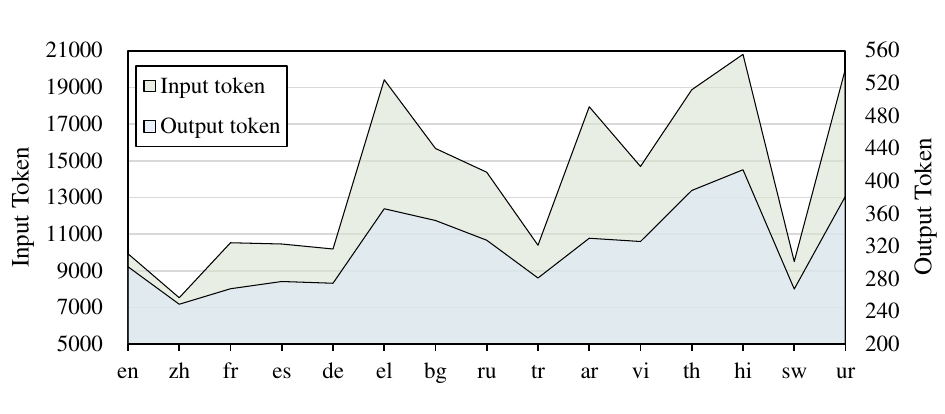}
    \caption{Statistics of average input token and output token for \textit{BaseAgent} method by GPT-3.5-turbo.}
    \label{fig:statistics_1}
\end{figure}

\subsection{Agentic system illustrates distinct multi- lingual token cost unfairness}
\label{sec:token}
Inspired by the inherent token-cost unfairness in multilingual LLMs~\citep{petrov2023language}, we investigate it by measuring the token consumption for both input and output in \modelname{}.
Interestingly, we identify that,
unlike traditional cases, where lower-resource languages face higher token costs, unfairness in agentic systems primarily stems from linguistic characters, not training resources.
As illustrated in Figure~\ref{fig:statistics_1}, most non-Latin script languages, especially on Greek (el) and Hindi (hi), consume as least twice as many tokens as english. Conversely,
languages like Chinese (zh) and German (de), which share Subject-Verb-Object (SVO) grammar structures, consume fewer tokens and mitigate concerns about token overuse.
These findings also emphasize the need and hopeful linguistic-related direction to address token-cost unfairness in multilingual agentic systems.

\subsection{Agentic system tends to overuse actions in languages with less speakers}
\label{sec:language_steps}
To assess the impact of multilingualism on action steps, we analyze the \textit{BaseAgent} methods of the top three performing models in Table~\ref{tab:step}.
The results reveal notable variations across languages, with less widely spoken languages, such as German (de), Turkish (tr), Vietnamese (vi), and Thai (th), requiring more steps.
For instance, in Qwen-2-7B-Instruct, the number of ``click'' actions for German and Turkish far exceeds the average of 3-4 clicks for correct interaction. Similarly, in GPT-series, Thai and Hindi exhibit disproportionately high ``search'' actions, far above the average of 1-2 searches for accurate outcomes. In contrast, Chinese, a widely spoken language\footnote{https://en.wikipedia.org/wiki/List\_of\_languages\_by\_\\total\_number\_of\_speakers}, demonstrates consistent performance and balanced action distributions across all models.
These findings indicate that languages with less speakers are more prone to overuse actions in \modelname{}. Therefore, addressing these action imbalances is essential for reducing global inequalities in agentic systems. 

\begin{figure}[t]
    \centering
    \includegraphics[width=1.0\columnwidth]{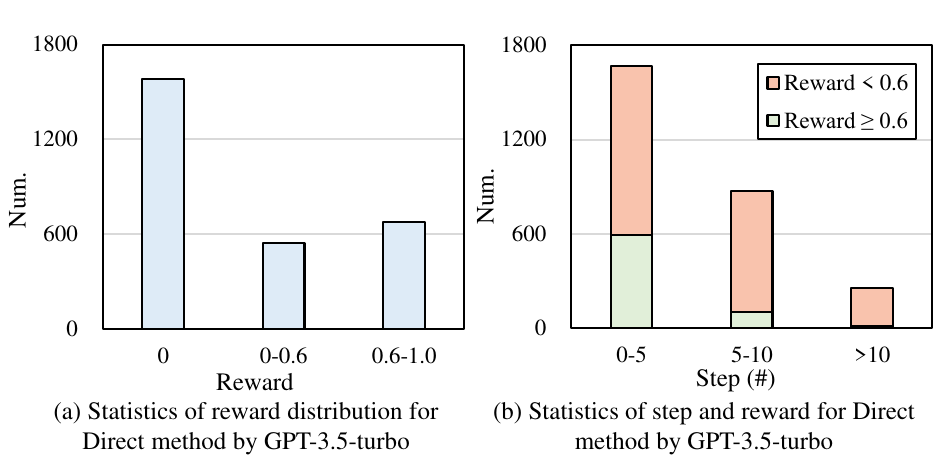}
    \caption{Statistics of action reward for \textit{BaseAgent} method by GPT-3.5-turbo.}
    \label{fig:statistics_2}
\end{figure}

\subsection{Agentic system performance poorly on long interaction in multilingual scenarios}
\label{sec:app:data_detailction-step}
As shown in Figure~\ref{fig:statistics_2} (a), GPT-3.5-turbo struggles to generate correct actions that yield sufficient rewards. To address this limitation, we notice that in pure English scenarios, the planning boundary of the model agent typically spans 5-10 steps, beyond which performance degrades~\citep{chen2024unlocking,hu2024hiagent,jin2024impact}.
Inspired by this, we analyze the reward distribution across all languages, as seen in Figure~\ref{fig:statistics_2} (b). Notably, most action rewards are concentrated within 0-5 steps, with a sharp decline thereafter\footnote{Human experts typically require more than 10 action steps in \modelname{}.}. This indicates that in multilingual scenarios, performance degradation occurs earlier than in English-based tasks.
It highlights that current language agents are limited to handling fewer steps and struggle with longer interactions in multilingual scenarios.
\begin{table}[t]
    \centering
    \begin{adjustbox}{width=\columnwidth}
      \begin{tabular}{lcccccc}
      \toprule
      Method & \cellcolor[rgb]{ .886,  .937,  .855}de & \cellcolor[rgb]{ .886,  .937,  .855}ru & \cellcolor[rgb]{ 1,  .949,  .8}tr & \cellcolor[rgb]{ 1,  .949,  .8}vi & \cellcolor[rgb]{ .988,  .894,  .839}bg & \cellcolor[rgb]{ .988,  .894,  .839}hi \\
      \midrule
      ReAct + En-ICL & 12.81  & 8.77  & 15.09  & 12.06  & 8.78  & 2.15  \\
      ReAct + De-ICL & 19.65  & 8.21  & 10.61 & 9.07  & 13.50  & 2.91 \\
      ReAct + L-ICL & 19.65 $\uparrow$ & 18.58 $\uparrow$ & 1.60 $\downarrow$ & 9.34 $\downarrow$ & 14.65 $\uparrow$ & 5.79 $\uparrow$ \\
      \bottomrule
      \end{tabular}%
    \end{adjustbox}
    \caption{Performance comparison of English, German, and local language ICL demonstration (En-ICL, De-ICL, and L-ICL) on GPT-3.5-turbo.``$\uparrow$'' represents the performance of L-ICL is better than En-ICL, and ``$\downarrow$'' indicates the opposite.}
    \label{tab:shot_lan}%
  \end{table}%

\section{Exploration}
Following \citet{qin2023cross}, we select 6 representative languages based on language statistics in CommonCrawl, including high- (de, ru), mid- (tr, vi), and low-resource (bg, hi) languages for exploration. Our study examines the effects of multilingual factors (\S~\ref{sec:shot-sample}, \S~\ref{sec:shot-number}, \S~\ref{sec:bolaa}) and agentic factors (\S~\ref{sec:translate-action}, \S~\ref{sec:reasoning-model}, and Appendix~\ref{sec:llm-scale}).

\subsection{Cross-lingual demonstrations can not introduce effective multilingual alignment}
\label{sec:shot-sample}

In-context Learning (ICL) with cross-lingual demonstrations has been shown to enhance performance, particularly for multilingual alignment~\citep{shi2023language}. Motivated by this, we evaluate ReAct~\cite{yao2023react} using demonstrations in English, German, and a local language (En-, De-, and L-ICL). In this framework, we adapt manually crafted ICL demonstrations to guide the model's next-step actions, with ``local language demonstration'' referring to cases where the ICL language matches the environment's language.
As shown in Table~\ref{tab:shot_lan}, we observe an unexpected result compared to traditional multilingual tasks~\citep{shi2023language, qin2024multilingual}. Specifically, performance of those with local language demonstrations exceeded that of cross-lingual demonstrations (En- and De-ICL) in most languages. This suggests that, unlike local language, current LLMs struggle to achieve robust multilingual alignment in ICL-based agentic scenarios.
This also highlights the need for future research to enhance cross-lingual transferability and better leverage multilingual capabilities in LLMs.

\begin{figure}[t]
    \centering
    \includegraphics[width=1.0\columnwidth]{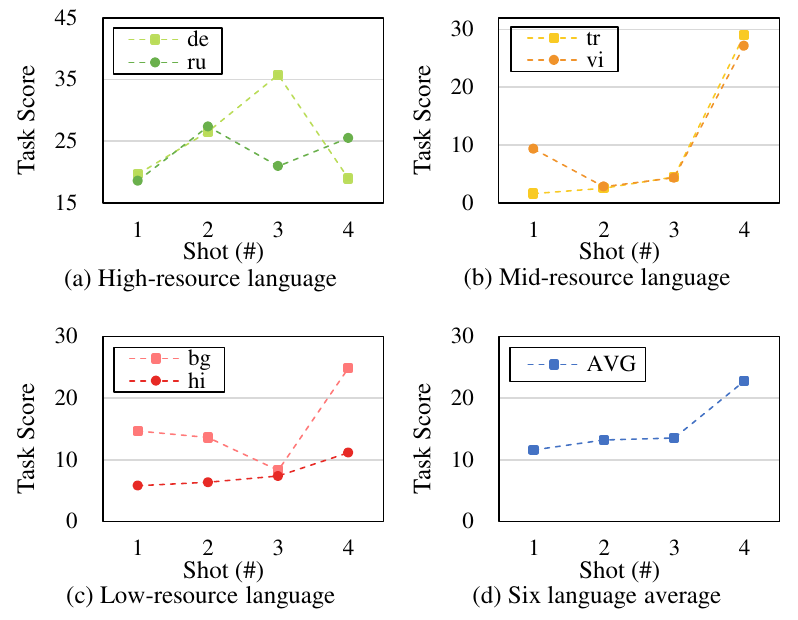}
    \caption{Performance of different shots in multilingual few-shot ReAct prompting.}
    \label{fig:shot_num}
\end{figure}
\subsection{The number of demonstrations has a positive effect on performance mainly in mid- and low-resource languages}
\label{sec:shot-number}
Further, inspired by~\citet{shi2023language,qin2024what},
we analyze how the number of demonstrations affects the performance of language agents within \modelname{}. The results yield the following insights:
\textit{\textbf{1. Limited benefits for high-resource languages:}} As shown in Figure~\ref{fig:shot_num} (a), task score initially improves with additional demonstrations but declines beyond a certain threshold.
\textit{\textbf{2. Substantial gains for mid- and low-resource languages:}} Figures~\ref{fig:shot_num} (b-d) shows that the task score improves steadily as the number of demonstrations increases.
It indicates that demonstration effectiveness depends on language resources. Mid- and low-resource languages require more demonstrations to boost their performance.

\subsection{Multilingual gaps are widespread in current English-centric agentic systems}
\label{sec:bolaa}
\begin{figure}[t]
    \centering
    \includegraphics[width=1.0\columnwidth]{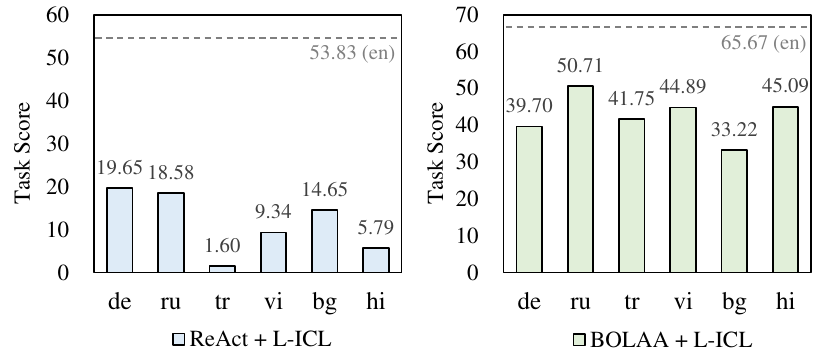}
    \caption{The performance of ReAct and BOLAA with L-ICL. Task score in English is from \citet{liu2024bolaa}.}
    \label{fig:bolaa}
\end{figure}

To comprehensively evaluate the multilingual limitation of current English-centric strong agentic frameworks in \modelname{}, we evaluated the multi-agent BOLAA~\cite{liu2024bolaa} in local language demonstration (BOLAA+L-ICL) using GPT-3.5-turbo and compared it with ReAct. The performance is shown in Figure~\ref{fig:bolaa}.
As observed, BOLAA+L-ICL performs well in \modelname{}, demonstrating the performance advantage of the multi-agent approach over the single-agent approach. But it still has a gap with English. This suggests that multilingual gaps are widespread in agentic systems. How to solve the understanding gap between different languages remains an important direction that needs to be explored in the future.

\subsection{LLMs overestimate their multilingual abilities rather than using translation tools}
\label{sec:translate-action}
Due to agents' high accuracy in English scenarios, an intuitive question arises: \textit{\textbf{Will agents autonomously use translation tools to overcome limitations in multilingual comprehension?}}
To explore this, we integrated a translation tool into the agent's action list, allowing it to translate web pages into English. Unfortunately, across all 14 languages, the LLMs never utilize the translation function, demonstrating overconfidence in its multilingual capabilities. This underscores the need for further research to enable agentic systems to intelligently utilize translation tools.

\subsection{The bottleneck of \modelname{} lies in language alignment, not complex logic}
\label{sec:reasoning-model}
\begin{figure}[t]
    \centering
    \includegraphics[width=1.0\columnwidth]{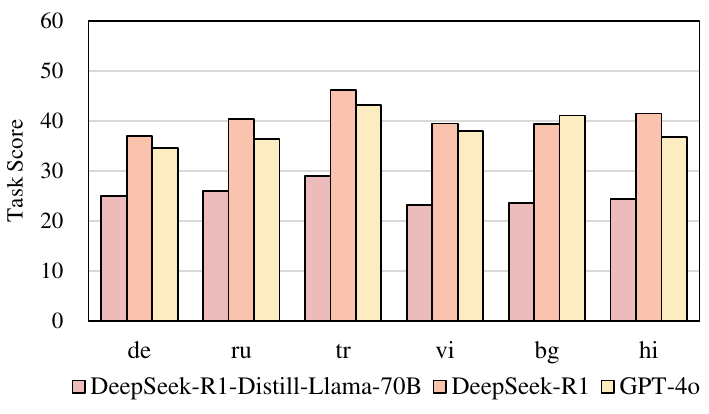}
    \caption{The performance comparison of reasoning models and GPT-4o.}
    \label{fig:reasoning}
\end{figure}

To examine whether the bottleneck of \modelname{} stems from complex reasoning, we evaluate DeepSeek-R1-Distill-Llama-70B and DeepSeek-R1~\cite{guo2025deepseek} using \textit{BaseAgent}. As shown in Figure~\ref{fig:reasoning}, the performance of reasoning models is lower than expected. Notably, DeepSeek-R1 performs similarly to GPT-4o, and the task score for all models are below 50.
This suggests that enhancing reasoning alone does not effectively support interaction in a multilingual interactive environment. Therefore, the main bottleneck of \modelname{} still lies in language alignment, rather than reasoning logic. Further research into multilingual alignment techniques is essential for LLMs' interaction capability improvements in multilingual environments.

\section{Related Work}

To evaluate the interaction capabilities of LLMs, researchers focus on benchmarks in interactive environments~\cite{toyama2021androidenv,sun2022meta,xie2024osworld}. Earlier, MiniWoB++~\cite{liu2018reinforcement} proposes a suite of web-browser tasks with over 100 web interaction environments. WebSRC~\cite{chen2021websrc} further integrates web pages and instructions for structural reading comprehension evaluation.
AndroidEnv~\cite{toyama2021androidenv} define custom tasks on the Android Operating System, enabling realistic simulation of an Android device.
After that, WebShop~\cite{yao2022webshop} emphasizes interactive tasks based on given instructions in e-commerce environments. Furthermore, Mind2Web~\cite{deng2024mind2web} extends agentic evaluation to 2,000 tasks across 137 websites. WebArena~\cite{zhou2024webarena} constructs a highly realistic and reproducible environment for language agents, which aims to boost the development of robust agents.

However, current interactive environments are primarily in English, which limits the development of multilingual agents. We are the first attempt to construct a multilingual interactive benchmark, \modelname{}, which quantifies the interaction performance of language agents.
\section{Conclusion}

To address the multilingual evaluation of language agents, we introduce \modelname{}, a benchmark comprising 14 languages, 2,800 text instructions, and 589,946 products. Through extensive experimentation on \modelname{}, we derive several novel insights into multilingual language agentic systens. We hope that \modelname{} can serve as an effective benchmark, providing valuable reference for future research.

\section*{Limitations}

We introduce a multilingual interactive web benchmark named \modelname{}, which explores the performance of understanding and interaction in the multilingual scenarios for language agents. Using Google Translate and multiple quality checks can only guarantee the accuracy and reliability of multilingual expressions as much as possible. However, the quality cannot be completely equivalent to that of experienced translators. In the future, we will explore more advanced data construction methods to achieve a more realistic multilingual environment.

\section*{Acknowledgments}

This work was supported by the National Natural Science Foundation of China (NSFC) via grant 62306342. This work was sponsored by the Excellent Young Scientists Fund in HunanProvince (2024JJ4070), the Science and Technology Innovation Program of Hunan Province under Grant 2024RC3024, and CCF-Zhipu Large Model Innovation Fund (NO.CCF-Zhipu202406). This work was also supported by the Key Laboratory of Data Intelligence and Advanced Computing in Provincial Universities, Soochow University. We are grateful for resources from the High Performance Computing Center of Central South University.
\bibliography{custom}

\clearpage
\appendix
\section*{Appendix}
\label{sec:app:data_detailppendix}

\section{Data Preparation Details}
\label{sec:app:data_detail}
Our instructions come from the test items in ReAct. The test instructions of ReAct are randomly selected instructions from the original WebShop. Then we find the corresponding 500 products according to the instructions. Based on the current products, we extract products with the same category from all products in WebShop as the product set of \modelname{}.

Language selection in \modelname{} is detailed in Table~\ref{tab:languages}: We select 14 languages covering a range of language families such as the Indo-European, Sino-Tibetan, and Afro-Asiatic language families, etc., including Chinese (zh), French (fr), Spanish (es), German (de), Greek (el), Bulgarian (bg), Russian (ru), Turkish (tr), Arabic (ar), Vietnamese (vi), Thai (th), Hindi (hi), Swahili (sw), and Urdu (ur).

\section{Manual Check and Re-Check Details}
\label{sec:app:check}
To ensure the translation quality, for every language, we extract 50 samples from the product data and instruction data each. The manual check involves the accuracy of the translation and appropriate expression with the help of well-trained translators.

As for the re-check process, we extract 20 samples for each language from the 200 instructions scored by GPT-4-turbo as mentioned below. All samples are reviewed by at least two well-trained individuals, and a kappa coefficient of 0.8 indicates the reliability of the process~\cite{landis1977measurement}.

\begin{table}[t]
    \centering
    \resizebox{\columnwidth}{!}{ 
      \begin{tabular}{ccc}
      \toprule
      Language Family & Language & Abbreviation \\
      \midrule
      \multirow{8}[1]{*}{Indo-European} & French & fr \\
            & Spanish & es \\
            & German & de \\
            & Greek & el \\
            & Bulgarian & bg \\
            & Russian & ru \\
            & Hindi & hi \\
            & Urdu  & ur \\
      \midrule
      Sino-Tibetan & Chinese & zh \\
      \midrule
      Turkic & Turkish & tr \\
      \midrule
      Afro-Asiatic & Arabic & ar \\
      \midrule
      Austroasiatic & Vietnamese & vi \\
      \midrule
      Tai-Kaidai & Thai  & th \\
      \midrule
      Niger-Congo & Swahili & sw \\
      \bottomrule
      \end{tabular}%
    }
    \caption{Language families and abbreviations mapping for the 14 \modelname{} languages}
    \label{tab:languages}%
  \end{table}%

For each check, the guideline instructions are as follows:
\begin{mybox}
  \textbf{Task:}Ensure that the entries translated from English into various languages are accurate and complete.
  \\ \hspace*{\fill} \\
  \textbf{[Instruction1]:} Open the translated sample in the target language.
  
  \textbf{[Instruction2]:} Identify and extract the content of the``instruction" field
  and the ``ASIN" (Amazon Standard Identification Number) from the translated sample.
  
  \textbf{[Instruction3]:} Locate the corresponding entry in the English dataset using the extracted ``ASIN."
  
  \textbf{[Instruction4]:} Compare the ``instruction" field from the translated sample with the ``instruction" field in the English dataset.
  
  \textbf{[Instruction5]:} Evaluate the translation for accuracy and completeness:
  \begin{itemize}
  \item
  If the translation is accurate and all essential elements are present, the entry is considered correct.
  \item
  If the translation is inaccurate or incomplete, record the ``ASIN" and the translation language for further review.
  \end{itemize}
\end{mybox}

\section{LLM Check Details}
\label{sec:app:llm_check}
We use GPT-4-turbo to score the translation quality. Specifically, we use the following prompt for scoring:
\begin{mybox}
    Original instruction: \dots

    Translated instruction: \dots

    Please rate the quality of the translation. Your answer should be a single number instead of any words. 
    
    The highest score is 10 and the lowest is 0.
\end{mybox}
After scoring, we build up the instruction set using those whose scores are higher than 8.0, with the scale of 200. In fact, except for Swahili and Vietnamese, every instruction in each language has reached the full score. Instructions in majority in Swahili and Vietnamese also reach the full score. Only fewer than 15\% get 9 or 8 points.

\section{Details in \modelname{}}
\label{sec:app:benchmark_detail}
\begin{figure}[t]
    \centering
    \includegraphics[width=1.0\columnwidth]{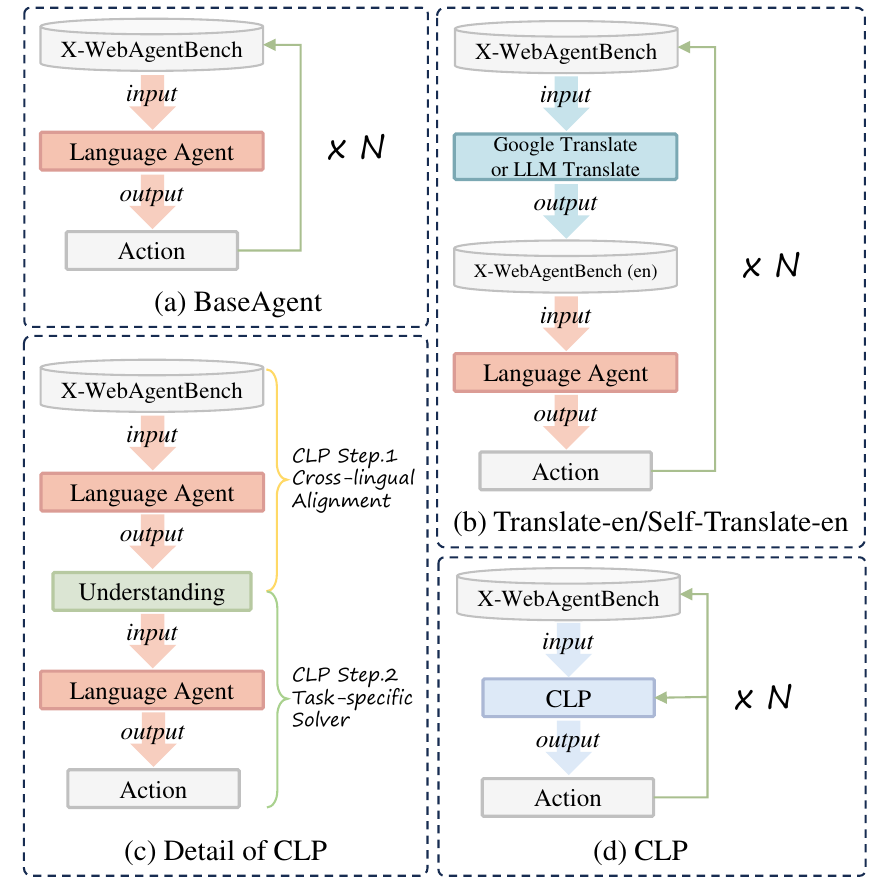}
    \caption{The diagram of these baseline methods. Figure (c) shows the multilingual alignment detail of \textit{CLP}. Language agent in \textit{BaseAgent}, \textit{Translate-en}, and \textit{Self-Translate-en} can read full history when they make a decision. Language agent in \textit{CLP} could not read history, but we input the previous action to the \textit{CLP} next decision as reference.}
    \label{fig:baseline}
\end{figure}

The actions within \modelname{} are detailed in Table~\ref{tab:action}. We use \textit{BaseAgent}, \textit{Translate-en}, and \textit{Self-Translate-en}, and \textit{CLP} as baselines to evaluate different LLMs. The detail of these methods is shown in Figure~\ref{fig:baseline}.

\begin{table}[t]
    \centering
    \begin{adjustbox}{width=\columnwidth}
    \begin{tabular}{lll}
      \toprule
      Type  & Action & Function \\
      \midrule
      search & search[keyword] & Search item by keyword \\
      click & click[prev/next page] & Jump to the prev/next item page \\
      click & click[back btu] & Back to search page \\
      click & click[item id] & Jump to the item page \\
      click & click[attribute] & View the attribute of the item \\
      click & click[buy] & Buy this item \\
      \bottomrule
      \end{tabular}%
    \end{adjustbox}
    \caption{Action list in the \modelname{}}
    \label{tab:action}%
  \end{table}%

\section{Prompt Settings and Format}
\label{sec:app:prompt}
We use the following prompt format for \modelname{} based on AgentBench, and $click$ and $search$ is multilingual prompt:
\begin{mybox}
    \textbf{[Instruction]}
    
    You are web shopping. I will give you instructions about what to do.
    You have to follow the instructions.
    \\ \hspace*{\fill} \\
    1.Every round I will give you an observation and a list of available actions, you have to respond an action based on the state and instruction. 
    \\ \hspace*{\fill} \\
    2.You can use search action if $search$ is available.
    You can click one of the buttons in clickable. An action should be of the following structure:
    \begin{itemize}
    \item
    $search$[keyword]: Search keywords on the search page. If you are not on the search page, please return to the search page. Such as ``$search$[$item$]".
    \item
    $click$[$prev$]/$click$[$next$]: Jump to the previous/next page. Such as ``$click$[$prev$]".
    \item
    $click$[$back$]: Back to search page. Such as ``$click$[$back$]".
    \item
    $click$[item id]: Click on an item to go to its details page. Such as ``$click$[ABCDE12345]".
    \item
    $click$[attribute]: Select item attribute on the details page. Such as ``$click$[$attr$]".
    \item
    $click$[$buy$]: Buy this item. Such as ``$click$[$buy$]".
    \end{itemize}
    3.If the action is not valid, perform nothing. 
    \\ \hspace*{\fill} \\
    4.Keywords in search are up to you, but the value in click must be a value in the list of available actions. 
    Remember that your keywords in search should be carefully designed. 
    \\ \hspace*{\fill} \\
    \textbf{[Response Format]}

    Thought: $<I\ think ...>$

    Action: $<$click$[something]>$
\end{mybox}

For some models that cannot follow the instructions, we use json to specify the output format:
\begin{mybox}
    Your response should use the following json format:
    
    \{``Thought":``I think ...", ``Action": ``$click$[something]"\}
\end{mybox}

For CLP, we use these prompt to align language firstly:
\begin{mybox}
    You are an expert in multi-lingual understanding. Please fully understand given observation and give the correct Action. 

    Please act as an expert in multi-lingual understanding in $language$.

    You need fully understand every words, and explain its effection.

    Don't response the Action, it is next step task.
    Let's understand the observation in English step-by-step!
\end{mybox}

Then, we use these prompts to get last action:
\begin{mybox}
    After understanding, you should act as an expert in reasoning in English. In the end, you should response Action in $language$ according previously understood observation.
    \\ \hspace*{\fill} \\
    Previous Action: $<prev\ action>$
    \\ \hspace*{\fill} \\
    You are web shopping. I will give you instructions about what to do.
    You have to follow the follow instructions.
    \\ \hspace*{\fill} \\
    1.Every round I will give you an observation and a list of available actions, you have to respond an action based on the state and instruction. 
    \\ \hspace*{\fill} \\
    2.You can use search action if search is available.
    You can click one of the buttons in clickable. An action should be of the following structure:
    \begin{itemize}
    \item
    $search$[keyword]: Search keywords on the search page. If you are not on the search page, please return to the search page. Such as ``$search$[$item$]".
    \item
    $click$[$prev$]/$click$[$next$]: Jump to the previous/next page. Such as ``$click$[$prev$]".
    \item
    $click$[$back$]: Back to search page. Such as ``$click$[$back$]".
    \item
    $click$[item id]: Click on an item to go to its details page. Such as ``$click$[ABCDE12345]".
    \item
    $click$[attribute]: Select item attribute on the details page. Such as ``$click$[$attr$]".
    \item
    $click$[$buy$]: Buy this item. Such as ``$click$[$buy$]".
    \end{itemize}
    3.If the action is not valid, perform nothing. 
    \\ \hspace*{\fill} \\
    4.Keywords in search are up to you, but the value in click must be a value in the list of available actions. 
    Remember that your keywords in search should be carefully designed. 
    \\ \hspace*{\fill} \\
    \textbf{[Response Format]}

    Thought: $<I\ think ...>$

    Action: $<$click$[something]>$
\end{mybox}

\begin{figure*}[t]
    \centering
    \includegraphics[width=1.0\textwidth]{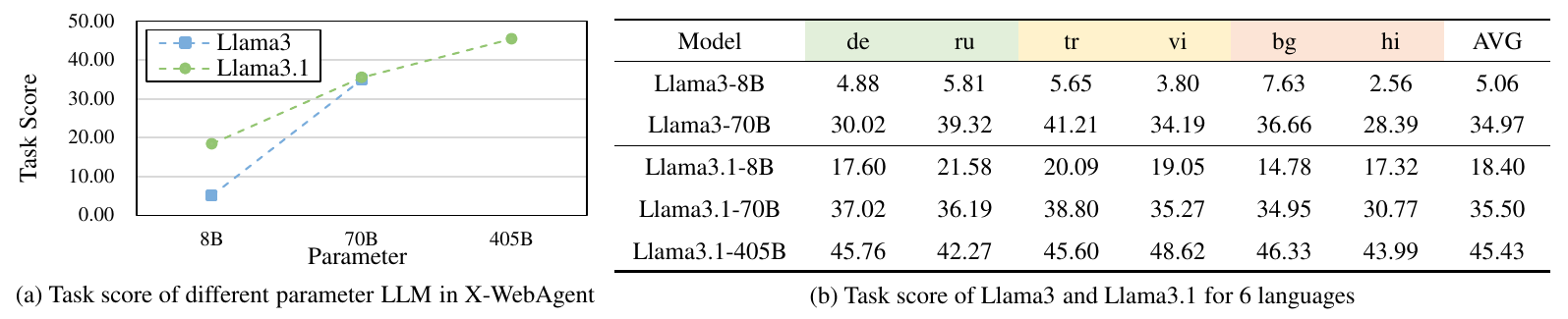}
    \caption{Performance comparison of different-scale LLMs on \modelname{}. Subfigure (a) presents the average performance of Llama3 and Llama3.1, while subfigure (b) provides detailed performance across 6 languages.}
    \label{fig:scale}
\end{figure*}

\begin{table*}[t]
    \centering
    \begin{adjustbox}{width=\textwidth}
        \begin{tabular}{lccccccccccccccc}
            \toprule
            Model& zh& fr& es& de& el& bg& ru& tr& ar& vi& th& hi& sw& ur& AVG\\
            \midrule
            DeepSeek-R1-Distill-Llama-70B& 18.22& 30.76& 21.84& 25.08& 22.84& 23.66& 25.95& 29& 21.58& 23.22& 17.44& 24.48& 21.55& 18.56& 23.16\\
            DeepSeek-R1& 34.84& 45.28& 43.2& 37.02& 38.61& 39.4& 40.44& 46.25& 25.51& 39.52& 20.47& 41.5& 26.64& 32.9& 36.54\\
            \bottomrule
            \end{tabular}%
    \end{adjustbox}
    \caption{The performance of DeepSeek-R1-Distill-Llama-70B and DeepSeek-R1 in \modelname{}.}
    \label{tab:rllm}%
  \end{table*}%

\begin{figure}[t]
    \centering
    \includegraphics[width=0.95\columnwidth]{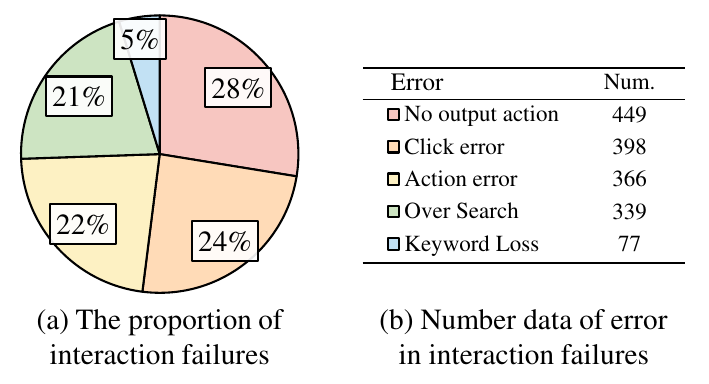}
    \caption{Statistics of action step (a), reward (b), and interaction failures (c and d). The output of the \textit{BaseAgent} method is from GPT-3.5-turbo.}
    \label{fig:error}
\end{figure}

\section{Multilingual performance follows LLM parameter increase}
\label{sec:llm-scale}

To assess the influence of LLM parameter scale on performance, we evaluate six languages representing high-, mid-, and low-resource groups.
As shown in Figure~\ref{fig:scale} (a, b), increasing LLM scale consistently improves both task-specific and average scores across all languages, though performance remains limited. These findings highlight a positive correlation between LLM scale and multilingual performance. However, despite enhancing multilingual capabilities, larger LLMs still struggle to address current multilingual challenges effectively.

\section{Performance of Reasoning Models}
We employ the Direct method to evaluate two reasoning large language models, DeepSeek-R1-Distill-Llama-70B and DeepSeek-R1. The final results are presented in Table~\ref{tab:rllm}.

\section{Error analysis of existing agentic system}
\label{sec:app:error}

To understand the factors resulting in language agent errors during interactions, we studied all the error cases in the \textit{BaseAgent} method outputs for 14 languages from the GPT-3.5-turbo logs. The final results are illustrated in Figure~\ref{fig:error}.

The main errors leading to interaction failures include the following five types: (1) \textit{No output action:} The output does not contain an action, preventing the next step from being executed; (2) \textit{Click error:} The button to be clicked is not on the page, rendering the instruction impossible; (3) \textit{Action error:} Use nonexistent commands on the web page, such as requesting a search action on a product page; (4) \textit{Over search:} Search times exceed the maximum limit; (5) \textit{Keywords loss:} Keywords in the instruction are not extracted.

Therefore, to enhance the decision-making capabilities of language agents on \modelname{}, the following abilities matter:
(1) Strengthen their understanding abilities in multilingual scenarios, which ultimately output correct actions.
(2) Reasonably plan the next instruction, avoiding repetitive and ineffective actions.

\section{Ethical Considerations}
\label{sec:app:ethical_considerations}
\paragraph{Data Access}
We obtained our data from the Webshop, which is open-source and freely accessible for academic research, which is in line with our commitment to ethical data usage.

\paragraph{Participant Recruitment}
We recruit participants from universities, ensuring that all have multilingual abilities. All annotators provided informed consent and were compensated above the local minimum wage. Additionally, the site does not require IRB review.

\paragraph{Dataset Check and Re-check Process}
Our checking process started with an onboarding test that introduced the task through 50 example questions. Participants were paid \$20 for this initial phase, which was designed to familiarize them with the task. Subsequently, annotators were paid \$15 per hour, averaging approximately 4 human annotation hours per person. In total, 42 experts participated in completing the annotation and recheck tasks.

\end{document}